%% file: main.tex
\documentclass[10pt,twocolumn,letterpaper]{article}

\usepackage{iccv}      


\definecolor{iccvblue}{rgb}{0.21,0.49,0.74}
\definecolor{ourblue}{rgb}{0.75,0.86,0.92}
\definecolor{ourred}{rgb}{1,0.66,0.71}
\usepackage[pagebackref,breaklinks,colorlinks,allcolors=iccvblue]{hyperref}
\usepackage{multirow,multicol,colortbl,makecell}
\usepackage[table]{xcolor}
\usepackage{tabularx}
\usepackage{siunitx}
\usepackage{adjustbox}
\usepackage[para]{footmisc}
\def\paperID{3888} 
\def\confName{ICCV}
\def\confYear{2025}

\title{Augmenting Moment Retrieval: Zero-Dependency Two-Stage Learning}

\author{
\textbf{Zhengxuan Wei}$^1$\textsuperscript{*} \quad
\textbf{Jiajin Tang}$^1$\textsuperscript{*} \quad
\textbf{Sibei Yang}$^2$\textsuperscript{\dag} \\
$^1$ShanghaiTech University \quad
{$^2$School of Computer Science and Engineering, Sun Yat-sen University}  \\
\texttt{\normalsize{\{weizhx2022, tangjj\}@shanghaitech.edu.cn}}\hspace{0.5cm}\texttt{\normalsize{yangsb3@mail.sysu.edu.cn}}\\
}

\begin{document}
\renewcommand{\thefootnote}{\fnsymbol{footnote}}
\maketitle

\setcounter{footnote}{0}
\footnotetext{
  $^{*}$ Equal contribution. \quad
  $^{\dagger}$ Corresponding author is Sibei Yang.
}

\input{sec/0_abstract}    
\input{sec/1_intro}
\input{sec/2_related_work}
\input{sec/3_method}
\input{sec/4_experiment}
\input{sec/5_conclusion}

{
    \small
    \bibliographystyle{ieeenat_fullname}
    \bibliography{main}
}

\end{document}

%% file: sec/0_abstract.tex
\begin{abstract}
Existing Moment Retrieval methods face three critical bottlenecks: (1) data scarcity forces models into shallow keyword-feature associations; (2) boundary ambiguity in transition regions between adjacent events; (3) insufficient discrimination of fine-grained semantics (e.g., distinguishing ``kicking" vs. ``throwing" a ball). In this paper, we propose a zero-external-dependency Augmented Moment Retrieval framework, AMR, designed to overcome local optima caused by insufficient data annotations and the lack of robust boundary and semantic discrimination capabilities. AMR is built upon two key insights: (1) it resolves ambiguous boundary information and semantic confusion in existing annotations without additional data (avoiding costly manual labeling), and (2) it preserves boundary and semantic discriminative capabilities enhanced by training while generalizing to real-world scenarios, significantly improving performance. Furthermore, we propose a two-stage training framework with cold-start and distillation adaptation. The cold-start stage employs curriculum learning on augmented data to build foundational boundary/semantic awareness. The distillation stage introduces dual query sets: Original Queries maintain DETR-based localization using frozen Base Queries from the cold-start model, while Active Queries dynamically adapt to real-data distributions. A cross-stage distillation loss enforces consistency between Original and Base Queries, preventing knowledge forgetting while enabling real-world generalization. Experiments on multiple benchmarks show that AMR achieves improved performance over prior state-of-the-art approaches. Code is available at \url{https://github.com/SooLab/AMR}.
\end{abstract}

%% file: sec/1_intro.tex
\section{Introduction}
\label{sec:intro}

The explosive growth of video platforms has spurred an urgent demand for the structured parsing of video content~\cite{wu2016harnessing,lin2019tsm,lu2016optasia}. As one of the core tasks in video semantic understanding~\cite{buch2022revisiting,sadhu2021visual,zeng2020dense,nan2021interventional}, moment retrieval~\cite{moon2023cgdetr,moon2023qd-detr,lei2021detecting,sun2024tr} aims to locate precisely the start and end times of target segments in long video streams based on textual descriptions. Although this task demonstrates significant value in intelligent search~\cite{korf1999artificial} and video summarization~\cite{wactlar2000informedia,zhang2016context,sharghi2016query}, its development has been consistently constrained by the scarcity of annotated data and annotation ambiguity. Dense annotation of a single video segment requires substantial manual effort~\cite{bianco2015interactive,vondrick2013efficiently,galasso2013unified,voigtlaender2021reducing}, while the blurred boundaries between adjacent events (such as ``preparing to shoot" and ``completing the shot") further exacerbate annotation inconsistency. Current mainstream methods often rely on additional data and pre-trained models to alleviate the data bottleneck, but these strategies typically necessitate the introduction of external knowledge or complex computational frameworks, leading to an imbalance between model generalization capability and resource overhead.
\begin{figure}
    \centering
    \includegraphics[width=0.9\linewidth]{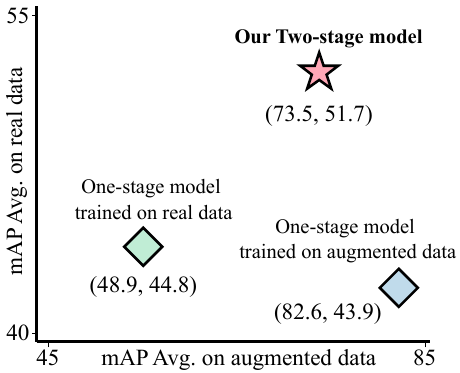}
    \caption{Traditional approaches (green diamonds) trained on real data show limited performance due to data scarcity. Data augmentation models (blue diamonds) often overfit to augmented distributions, degrading real data performance. Our two-stage training strategy (red pentagram) effectively leverages augmented data while markedly improving real data performance.}
    \label{fig:intro}
    \vspace{-4mm}
\end{figure}

\begin{figure*}
    \centering
    \includegraphics[width=1\linewidth]{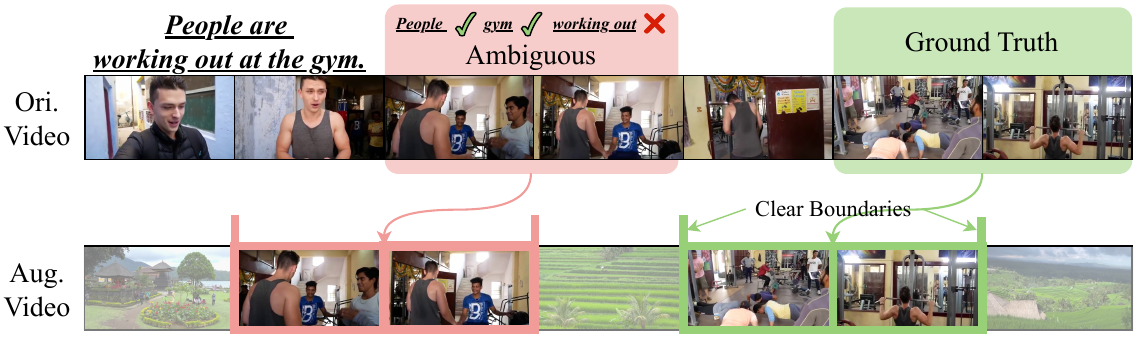}
    \caption{Overview of our Splice-and-Boost augmentation strategy. The Splice operation combines foreground segments from one video with background intervals of another to create samples with clear boundaries. Simultaneously, the Boost mechanism introduces semantically ambiguous segments into these spliced samples. This helps enhance the model's ability to distinguish between similar events.}
    \label{fig:aug}
    \vspace{-4mm}
\end{figure*}

Existing research primarily advances along three paths: feature alignment-based architecture optimization methods~\cite{yang2024taskweave,sun2024tr} enhance local feature alignment by designing multimodal interaction modules, yet their constructed attention mechanisms exhibit limited discriminative power for semantically similar segments; pre-training-based knowledge transfer methods (e.g., SG-DETR~\cite{gordeev2024saliency}) leverage large-scale vision-language models to improve semantic representation, while external large-model-based data synthesis method~\cite{paul2024videolights} generate virtual video-text pairs, but the billion-level parameter counts result in skyrocketing deployment costs. These three approaches have failed to address a core contradiction: \textbf{how to mine discriminative spatiotemporal reasoning patterns from limited annotated data under zero external dependencies?}

Through in-depth analysis of existing paradigms, we have identified three key challenges: \textbf{First, data sparsity leads models to fall into local optima.} When training samples are insufficient, models tend to establish shallow associations between textual keywords and local video features rather than understanding the temporal integrity of actions. \textbf{Second, boundary ambiguity undermines localization robustness.} The discrimination of transition regions between adjacent events (e.g., overlapping frames between ``preparing to shoot" and ``completing the shot") lacks clear boundary signals. These phenomena essentially stem from the insufficient mining of internal discriminative information in traditional methods. \textbf{Third, similar semantic discrimination capability is inadequate.} Existing methods lack effective mechanisms for distinguishing between semantically similar but logically or detail-wise different segments (e.g., ``kicking a ball" and ``throwing a ball" both involve ball movements).

To address these challenges, this paper proposes two core innovations: \textbf{(1) Splice and Boost data augmentation strategy} that stimulates model boundary and semantic discrimination capabilities through structured recombination; (2) \textbf{Two-stage training framework with cold-start and distillation } that facilitates knowledge transfer between augmented data and real-world scenarios. For Splice and Boost strategy, we design a dynamic video splicing mechanism: extracting target segments, semantically similar interference segments, and background segments from different videos, then generating enhanced samples with clear decision boundaries and interference samples with similar semantics through their combination. 

However, due to the inherent limitations of current single-stage training frameworks in effectively leveraging the combined benefits of augmented and real data, these models tend to converge to trivial solutions that merely capture mutation boundary characteristics from the augmented data. As shown in Figure~\ref{fig:intro}, current single-stage training model (\textcolor{ourblue}{\textbf{blue}}) exhibits significant performance limitations when applied to real-world data. To address this and achieve effective transfer of augmented capability, we construct a two-stage progressive training pipeline with a query-based transformer architecture~\cite{um2025kwdetr, sun2024trdetr, carion2020detr, moon2023qd-detr, zhu2025rethinking,chen2024survey12,huang2023free38, huang2025mvtokenflow39, shi2023edadet62, shi2023logoprompt63, shi2024plain66, tang2023contrastive70, zheng2023ddcot88}. The cold-start stage adopts a curriculum learning strategy, using augmented data to establish fundamental boundary and semantic discrimination capabilities in the model; The distillation adaptation stage innovatively designs a dual-path distillation mechanism with two distinct query sets, each with clearly defined roles: the \textbf{original query} set is designed to maintain boundary localization capability within the DETR architecture, while the \textbf{active query} set is designed to dynamically absorb real data characteristics. To prevent knowledge forgetting, cross-stage distillation loss functions are employed to constrain the distribution consistency between the \textbf{original query} and the \textbf{base query}, where the \textbf{base query} is trained from the cold-start model. This approach ensures a seamless transition and knowledge retention across stages, enabling the model to effectively adapt to real-world data while preserving its foundational capabilities learned from the cold-start stage (Figure~\ref{fig:intro} \textcolor{ourred}{\textbf{red}}).

\noindent The contributions of this paper are multi-fold: 
\begin{itemize}
    \item Proposing a zero-external-dependency video splicing augmentation theory that breaks through the dual constraints of data scarcity and distribution bias through a decoupling-recombination strategy;
    \item Designing a stable transfer mechanism based on dual-path distillation, solving the knowledge compatibility issue between augmented data and real-world scenarios.
    \item Our method demonstrates consistent and significant performance improvements across all moment retrieval benchmarks, outperforming all existing state-of-the-art approaches.
\end{itemize}

%% file: sec/2_related_work.tex
\section{Related Work}
\label{sec:related_work}

\textbf{Moment Retrieval (MR)} is a multi-modal grounding task~\cite{dai2024curriculum20, lin2021structured46, shi2024part2object64, tang2023temporal71, yang2021bottom81, hu2016segmentation}, aiming to localize video segments corresponding to textual queries. Early approaches fall into two categories: proposal-based methods~\cite{gao2017ctrl, anne2017mcn, xu2019qspn, chen2019sap, chen2018tgn, zhang2020tan, liu2024tuning}, which rely on pre-generated candidate segments, and proposal-free methods~\cite{chen2020gdp, lu2019debug, zeng2020drn, ghosh2019excl, zhang2020vslnet, rodriguez2020proposal, rodriguez2021dori}, which directly predict temporal boundaries without proposals. Recent advancements leverage DETR-based architectures~\cite{lei2021qvhighlights,moon2023qd-detr,sun2024trdetr,jang2023eatr,lee2025bamdetr,xiao2024uvcom,um2025kwdetr,tang2025simdetr,tang2023cotdet,zhang2025temporally}, inspired by their success in object detection, to improve MR performance through end-to-end query modeling. Concurrently, multimodal large language models (MLLMs)~\cite{meinardus2024mrblip,lu2024llavamr,huang2024vtimellm,wang2024timerefine,huang2024lita,guo2024vtgllm} exploit cross-modal knowledge to achieve remarkable results by aligning linguistic and visual semantics.

\noindent Beyond architectural innovations, data-centric approaches have emerged to address MR challenges. Some works~\cite{lu2019debug, chen2020gdp, zeng2020drn, rodriguez2023memory} tackle data imbalance issues or expensive computational costs, while others incorporate large-scale external datasets for pretraining. Notably, VIDEOLIGHTS~\cite{paul2024videolights} leverages BLIP-2’s image-to-text generation to synthesize training data. Although effective, these methods often depend on external resources. In contrast, we propose a simpler, generalizable data augmentation strategy that enhances performance without external data or pretrained models.

\begin{figure*}
    \centering
    \includegraphics[width=1\linewidth]{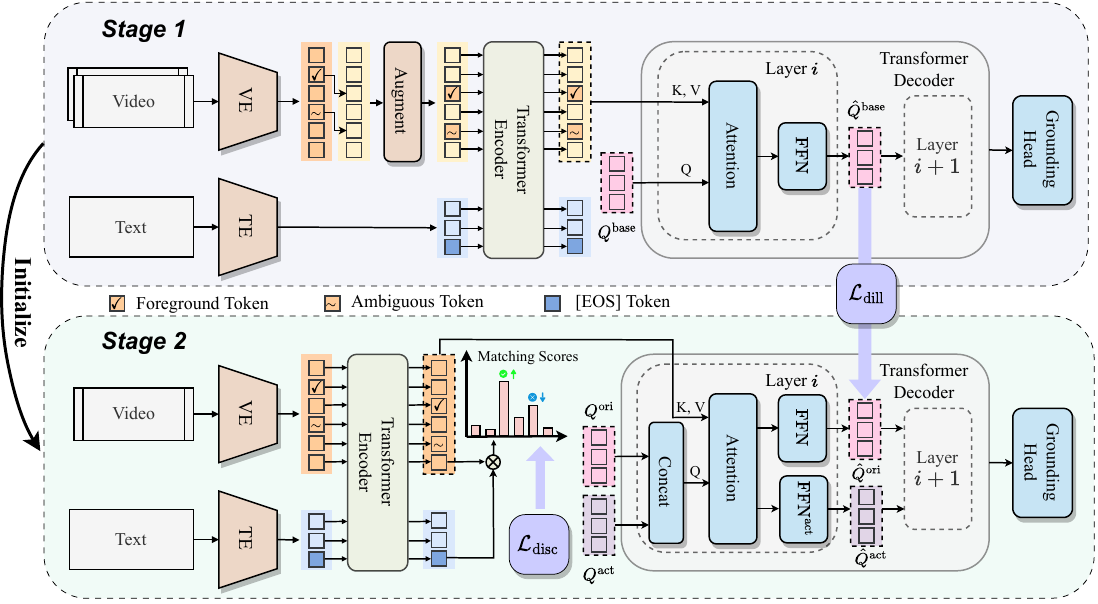}
    \caption{Overview of our proposed AMR framework. The training is conducted in two stages: In the first stage, we introduce a Splice-and-Boost augmentation strategy that synthesizes diverse training samples by splicing foreground segments into background videos and injecting ambiguous distractors. In the second stage, we use dual-path distillation to transfer the knowledge from the first stage, and a discriminative contrastive loss reinforces the distinction between true events and ambiguous fragments.}
    \label{fig:enter-label}
    \vspace{-4mm}
\end{figure*}

\noindent \textbf{Video Data Augmentation} has been widely explored to address the challenges of limited and difficult-to-annotate video data. Common augmentation techniques involve spatial transformations, such as random flipping and cropping, or temporal modifications, such as random downsampling. In recent years, novel augmentation strategies have emerged. For instance,~\cite{yun2020videomix} introduced a method for video classification that randomly mixes videos along spatial and temporal dimensions, assigning ground truth labels proportionally based on voxel distribution. Building on this,~\cite{zhang2020selfpaced} employed GAN~\cite{goodfellow2020gan} to generate more realistic video data, while~\cite{gowda2022learn2augment} proposed a learning-based approach to select optimal video pairs for higher-quality augmentation. Additionally,~\cite{pan2024labelefficient} utilized video diffusion models to generate labeled videos for the guidewire segmentation task.

\noindent However, these techniques are designed for classification or segmentation, leaving MR-specific augmentation underexplored. Our work bridges this gap with a task-oriented augmentation framework tailored to MR’s unique requirements, prioritizing temporal-textual alignment over generic spatial or temporal perturbations.

%% file: sec/3_method.tex
\section{Method}
\label{sec:method}

Existing moment retrieval frameworks (e.g., M-DETR~\cite{lei2021qvhighlights}) face significant challenges under data scarcity and distribution imbalance. While DETR-based architectures~\cite{sun2024trdetr, yang2024taskweave, xiao2024uvcom, lee2025bamdetr} achieve end-to-end localization through encoder-decoder paradigms, their performance is constrained by two critical limitations: 1) High annotation costs for video data result in limited training samples, causing models to overfit specific training distributions; 2) Practical scenarios contain inherently ambiguous event boundaries with adjacent segment overlaps and semantically similar events that induce localization confusion.

To address these limitations, we propose a novel data augmentation approach coupled with a dual-path distillation two-stage training framework AMR. This framework first employs a Splice and Boost strategy to enhance data diversity by explicitly constructing samples with well-defined boundaries and semantic distractors, followed by a two-stage training mechanism that resolves distribution discrepancies between augmented and real-world data through two-phase optimization. Requiring no external data or pretrained models, the framework enhances boundary awareness and semantic discrimination via synthetic data and subsequently transfers these capabilities through distillation.

Our work focuses on leveraging limited and imbalanced training samples to enhance temporal localization models' generalization capabilities in real-world scenarios. Consequently, the proposed AMR adheres to the standard vanilla DETR-based framework, comprising feature extraction, transformer encoder, and transformer decoder modules. For feature extraction, we align with previous work by using CLIP~\cite{radford2021clip} and SlowFast~\cite{feichtenhofer2019slowfast} to extract video features while employing the CLIP text encoder for text feature extraction, where the encoded video features are denoted as \(V \in \mathbb{R}^{L \times C}\) (\(L\) video frames) and the text features as \(T \in \mathbb{R}^{K \times C}\) (\(K\) text tokens), with \(C\) representing the hidden dimension. In the transformer encoder, cross-attention from visual to text features encodes multimodal video representation \(\hat{V} \in \mathbb{R}^{L \times C}\) and text representation \(\hat{T} \in \mathbb{R}^{K \times C}\). The decoder processes \(N\) learnable queries \(Q \in \mathbb{R}^{N \times C}\) with video features through several attention blocks and feed-forward networks, generating updated queries \(\hat{Q} \in \mathbb{R}^{N \times C}\). Final boundary coordinates and confidence scores are predicted through bounding box and classification heads.

\subsection{Data Augmentation: Splice and Boost}
This section introduces our Splice-and-Boost augmentation strategy to address limited and imbalanced training samples. The Splice operation synthesizes samples with clear boundaries by splicing foreground segments from one video with background intervals of another. Concurrently, the Boost mechanism explicitly injects semantically ambiguous segments into spliced samples to enhance the model’s discriminative capability against similar events. Note that, all augmentation operations are performed exclusively using existing dataset resources, ensuring no external data is introduced.

\noindent \textbf{Temporal Splicing with Boundary Enhancement  }
To mitigate data scarcity and generate data with well-defined boundaries, we first extract ground-truth video segments from raw samples and apply random undersampling to all segments to boost data diversity. Next, we choose another video as background material. For each undersampled target segment, we randomly identify a window of equal length on the background video's continuous timeline and delete it. Afterward, the processed target segment is seamlessly inserted into this blank space. This splicing mechanism strengthens boundary awareness by decoupling segment semantics from background dependencies, thereby enhancing the model’s ability to localize events independently of their surrounding context.

\noindent \textbf{Semantic Boosting via Hard Negative Mining}
To further improve semantic discrimination, the Boost mechanism introduces challenging ambiguous segments as hard negatives. These segments are identified through a cross-validation paradigm: the training set $\mathcal{D}_{\text{train}}$ is split into two mutually exclusive subsets $\mathcal{D}_1$ and $\mathcal{D}_2$ ($\mathcal{D}_1 \cup \mathcal{D}_2 = \mathcal{D}_{\text{train}}$, $|\mathcal{D}_1| = |\mathcal{D}_2|$). We first train a model on $\mathcal{D}_1$ and detect high-confidence false positives on $\mathcal{D}_2$. A predicted segment $W_i \in \mathbb{R}^{N \times 2}$ with confidence score $S_i \in \mathbb{R}^N$ is retained as an ambiguous segment if:  
(1) $\text{IoU}(W_i, W^{\text{gt}}_j) = 0$ for all $j \in \{1, 2, \ldots, N^{\text{gt}}\}$, where $W^{\text{gt}} \in \mathbb{R}^{N^{\text{gt}} \times 2}$ denotes ground-truth segments;  
(2) $S_i > \theta$, where $\theta$ is a predefined threshold.  
By swapping the roles of $\mathcal{D}_1$ and $\mathcal{D}_2$, we iteratively collect ambiguous segments across the entire $\mathcal{D}_{\text{train}}$.  

The full Splice and Boost pipeline proceeds as follows: (1) Extract ambiguous segments via the above cross-validation; (2) Randomly downsample both ground-truth and ambiguous segments; (3) Seamlessly splice these segments into blank intervals of a background video from an unrelated source. By jointly optimizing the accurate localization of true segments and suppression of ambiguous ones, our strategy balances diversity enhancement and discriminative capability under limited data without requiring external resources.  

\subsection{Two-stage Training: Cold Start and Distillation}
In traditional single-stage training frameworks~\cite{sun2024trdetr, yang2024taskweave, xiao2024uvcom, lee2025bamdetr}, the distribution discrepancy between augmented data \(\mathcal{D}_{\text{aug}}\) and real data \(\mathcal{D}_{\text{real}}\) often leads to a decline in model generalization. Specifically, the sharp boundaries of events in synthetic data may cause the model to over-rely on abrupt visual features (e.g., scene cuts or sudden object appearances) while neglecting gradual transition patterns in real-world scenarios (e.g., natural continuity of human actions). To address this issue, we propose a cold-start distillation two-stage training framework (Fig. 3), whose core idea is to establish fundamental boundary localization and semantic modeling capabilities through augmented data in the cold-start stage and achieve migration to real data via a knowledge distillation strategy in the distillation adaptation stage.

\subsubsection{Two-stage Training Framework}
During the cold-start stage, we train the model exclusively on augmented data, aiming to enhance its semantic understanding and robust segment localization capabilities. Since augmented data explicitly distinguishes foreground from background and includes challenging segments, this stage improves the model’s discriminative ability in complex scenes and its boundary localization accuracy. However, due to the distribution gap between augmented and real data, directly applying the model trained in the cold-start stage often yields suboptimal performance on real data. To address this limitation, we introduce a second stage of training, where we further refine the base model obtained from the cold-start stage using real data to gradually adapt it to the distribution of real-world scenarios.

\subsubsection{Dual-path Distillation}
 Yet, direct fine-tuning risks the model forgetting the discriminative capabilities learned from \(\mathcal{D}_{\text{aug}}\). To preserve the boundary awareness and semantic discriminative capabilities acquired during the cold-start phase, we develop a dual distillation mechanism that integrates discriminative contrastive loss and query knowledge preservation. This strategy addresses two critical challenges: maintaining the model’s ability to distinguish ambiguous fragments without over-suppressing semantically relevant background content and stabilizing the learned temporal localization patterns when adapting to real data distributions. Together, these mechanisms ensure robust adaptation while retaining essential capabilities.

Recognizing that temporal localization patterns are predominantly encoded in the decoder's query interactions, we devise a progressive adaptation strategy with parameter-isolated knowledge preservation. During the distillation stage, we introduce active queries \(Q^{\text{act}} \in \mathbb{R}^{N \times C}\) alongside the original queries \(Q^{\text{ori}}\), each paired with dedicated feed-forward networks (FFN\(^{\text{act}}\)). The decoder computation evolves into:  

\begin{equation}
    \begin{aligned}
Q^{\text{ori}}_{\text{att}}, Q^{\text{act}}_{\text{att}} = \text{CA}(\text{SA}([Q^{\text{ori}},Q^{\text{act}}]), \hat{V}, \hat{V}), \\ \hat{Q}^{\text{ori}} = \text{FFN}(Q^{\text{ori}}_{\text{att}} ),\hat{Q}^{\text{act}} = \text{FFN}^{\text{act}}(Q^{\text{act}}_{\text{att}}),
\end{aligned}
\end{equation}

To prevent catastrophic forgetting of cold-start acquired knowledge, we impose a distillation constraint between the base model and the original query pathway. The distillation loss measures feature consistency in the decoded query representations. Denote the decoded query in the base model as \(\hat{Q}^{\text{base}}\), \(\mathcal{L}_{\text{dill}}\) is computed as:
\begin{equation}
\mathcal{L}_{\text{dill}} = 1 - \frac{1}{N}\sum_{i=1}^{N}\frac{(\hat{Q}^{\text{ori}}_{i})^T \hat{Q}^{\text{base}}_{i}}{\|\hat{Q}^{\text{ori}}_{i}\|\|\hat{Q}^{\text{base}}_{i}\|}
\end{equation}

This loss is applied at each decoder layer, allowing the model to progressively integrate real-data patterns through active components while anchoring the established localization capabilities via original parameters. The dual-path architecture effectively bridges the synthetic-to-real domain gap by mediating knowledge transfer through parameter isolation and similarity-driven regularization. 

\subsubsection{Discriminative Contrastive Loss}
Additionally, we formulate a contrastive learning objective to enforce relative discriminability between ground-truth segments and ambiguous fragments. Given the encoded video features \(\hat{V} \in \mathbb{R}^{L \times C}\) and text representation \(\hat{T}_{\text{eos}} \in \mathbb{R}^C\) from the [EOS] token, we compute frame-level text-video matching scores \(M \in \mathbb{R}^L\) via cosine similarity:  

\begin{equation}
M_i = \frac{\hat{V}_i \hat{T}_{\text{eos}}^\top}{\|\hat{V}_i\| \|\hat{T}_{\text{eos}}\|}, \quad \forall i \in \{1,...,L\}
\end{equation}

Instead of imposing absolute score suppression on ambiguous regions — which risks penalizing legitimate semantic correlations — we construct a relative ranking constraint. Let \(p\) denote the average matching score within ground-truth segments \(W^{\text{gt}}\), and \(n\) represent the average score over ambiguous regions \(W^{\sim}\), the contrastive loss formula is:  

\begin{equation}
\begin{aligned}
p = &\frac{\sum_{i=1}^{L}\mathbb{I}_{W^{\text{gt}}}(i)M_i}{\sum_{i=1}^{L}\mathbb{I}_{W^{\text{gt}}}(i)}, n = \frac{\sum_{i=1}^{L}\mathbb{I}_{W^{\sim}}(i)M_i}{\sum_{i=1}^{L}\mathbb{I}_{W^{\sim}}(i)}, \\ &\mathcal{L}_{\text{disc}} = -\log\frac{\exp(p /\tau)}{\exp(p /\tau) + \exp(n /\tau)},
\end{aligned}
\end{equation}

where we define \(\mathbb{I}_{W}(i)\) as follows: \(\mathbb{I}_{W}(i)=1\) if there exists a segment \(w\) in \(W\) such that the \(i\)-th frame is within segment \(w\); otherwise, it is 0. And \(\tau\) serves as a temperature parameter controlling separation intensity. This formulation encourages the model to maintain higher discriminative margins between true events and their semantic neighbors without suppressing legitimate background correlations.

\subsection{Training Objective} 

This section details the optimization strategies and training objectives adopted in our framework.

\noindent\textbf{Hungarian Matching}
To establish one-to-one correspondence between predicted and ground-truth objects without manual heuristics, we employ the Hungarian matching algorithm following the DETR framework~\cite{carion2020detr}. The matching task is formulated as a minimum-cost bipartite graph matching problem. Specifically, a bipartite graph is constructed between the prediction set (containing \(N\) elements) and the padded ground-truth set (expanded to \(N\) elements with empty sets). The optimal assignment is determined by minimizing the matching cost:

\begin{equation}
\begin{aligned}
\mathcal{C}_{\text{match}} = \lambda_{\text{cls}}S_i + \lambda_{\text{loc}}\mathcal{L}_{\text{loc}}(W_i, W_j^{\text{gt}}),
\end{aligned}
\end{equation}

\noindent where \(S_i\) denotes the confidence score of predictions, \(\mathcal{L}_{\text{loc}}\) represents the localization loss combining L1 and GIoU losses, and \(\lambda_{\text{cls}}\) and \(\lambda_{\text{loc}}\) are weighting parameters. The Hungarian algorithm identifies the permutation with minimal total cost, enabling end-to-end training without post-processing steps such as Non-Maximum Suppression (NMS).  

\noindent\textbf{Loss Function}  
During the cold-start stage training, we utilize the baseline loss functions from M-DETR~\cite{lei2021qvhighlights}, including the classification loss \(\mathcal{L}_{\text{cls}}\), the localization loss \(\mathcal{L}_{\text{loc}}\), and the auxiliary saliency loss \(\mathcal{L}_{\text{sal}}\) (see \cite{lei2021qvhighlights} for implementation details). The composite loss function is defined as: 

\begin{equation}
\begin{aligned}
\mathcal{L}_{\text{cd}} = \lambda_{\text{cls}} \mathcal{L}_{\text{cls}} + \lambda_{\text{loc}} \mathcal{L}_{\text{loc}} + \lambda_{\text{sal}} \mathcal{L}_{\text{sal}},
\end{aligned}
\end{equation} 

\noindent where \(\lambda_{\text{cls}}\), \(\lambda_{\text{loc}}\), and \(\lambda_{\text{sal}}\) are weight hyperparameters.  

In the distillation stage training, we enhance the objective function by incorporating two proposed components: distillation loss \(\mathcal{L}_{\text{distill}}\) and discriminative contrastive loss \(\mathcal{L}_{\text{disc}}\). The updated loss function is formulated as:  

\begin{equation}
\begin{aligned}
\mathcal{L}_{\text{cd}} = \mathcal{L}_{\text{stage1}} + \lambda_{\text{dill}} \mathcal{L}_{\text{dill}} + \lambda_{\text{disc}} \mathcal{L}_{\text{disc}},
\end{aligned}
\end{equation} 

\noindent where \(\lambda_{\text{dill}}\) and \(\lambda_{\text{disc}}\) are weight hyperparameters.

\input{table/qvh}

%% file: table/qvh.tex
\begin{table*}[h]
\centering
\setlength\tabcolsep{2.6mm} 
\renewcommand\arraystretch{0.9}
\begin{tabular}{l|cc|ccc|cc|ccc}
\toprule
\multirow{4}{*}{\textbf{Method}} & \multicolumn{5}{c|}{\textbf{Validation}} & \multicolumn{5}{c}{\textbf{Test}} \\
\cmidrule(lr){2-6} \cmidrule(lr){7-11}
& \multicolumn{2}{c|}{R1} & \multicolumn{3}{c|}{mAP} & \multicolumn{2}{c|}{R1} & \multicolumn{3}{c}{mAP} \\
\cmidrule(lr){2-3} \cmidrule(lr){4-6} \cmidrule(lr){7-8} \cmidrule(lr){9-11}
& @0.5 & @0.7 & @0.5 & @0.75 & Avg. & @0.5 & @0.7 & @0.5 & @0.75 & Avg. \\ 
\midrule

M-DETR~\cite{lei2021qvhighlights} & 53.94 & 34.84 & - & - & 32.20 & 52.89 & 33.02 & 54.82 & 29.40 & 30.73 \\
UMT~\cite{liu2022umt} & 60.26 & 44.26 & - & - & 38.59 & 56.23 & 41.18 & 53.83 & 37.01 & 36.12 \\
QD-DETR~\cite{moon2023qd-detr} & 62.68 & 46.66 & 62.23 & 41.82 & 41.22 & 62.40 & 44.98 & 62.52 & 39.88 & 39.86 \\
UniVTG~\cite{lin2023univtg} & 59.74 & - & - & - & 36.13 & 58.86 & 40.86 & 57.60 & 35.59 & 35.47 \\
EaTR~\cite{jang2023eatr} & 61.36 & 45.79 & 61.86 & 41.91 & 41.74 & - & - & - & - & - \\
MomentDiff~\cite{li2023momentdiff} & - & - & - & - & - & 57.42 & 39.66 & 54.02 & 35.73 & 35.95 \\
TR-DETR~\cite{sun2024trdetr} & 67.10 & 51.48 & 66.27 & 46.42 & 45.09 & 64.66 & 48.96 & 63.98 & 43.73 & 42.62 \\
TaskWeave~\cite{yang2024taskweave} & 64.26 & 50.06 & 65.39 & 46.47 & 45.38 & - & - & - & - & - \\
UVCOM~\cite{xiao2024uvcom} & 65.10 & 51.81 & - & - & 45.79 & 63.55 & 47.47 & 63.37 & 42.67 & 43.18 \\
BAM-DETR~\cite{lee2025bamdetr} & 65.10 & 51.61 & 65.41 & 48.56 & 47.61 & 62.71 & 48.64 & 64.57 & 46.33 & 45.36 \\
\rowcolor{gray!50} 
\textbf{Ours} & \textbf{70.13} & \textbf{56.65} & \textbf{70.28} & \textbf{53.20} & \textbf{51.66} & \textbf{68.22} & \textbf{51.88} & \textbf{68.53} & \textbf{49.82} & \textbf{48.43} \\
\bottomrule
\end{tabular}
\caption{Experimental results on the QVHighlights validation and test set. \textbf{Bold} fonts indicate the best results.}
\label{tab:1}
\end{table*}

%% file: sec/4_experiment.tex
\section{Experiment}
\label{sec:experiment}
\subsection{Datasets and Metrics}
\textbf{Datasets} Our study evaluates performance on three benchmark datasets for moment retrieval: QVHighlights~\cite{lei2021qvhighlights}, Charades-STA~\cite{gao2017ctrl}, and TACoS~\cite{regneri2013tacos}. QVHighlights consists of 10,148 videos, primarily sourced from YouTube, including vlogs and news clips. Notably, this dataset allows a single textual query to correspond to multiple segments within a video, making it more representative of real-world applications. Therefore, we adopt it as our primary benchmark. Charades-STA comprises 9,848 videos, focusing on human activities in indoor scenes. TACoS consists of 127 cooking-related videos, characterized by long video durations and short target segments, posing a significant challenge for moment retrieval.

\input{table/tacos_and_cha}

\noindent\textbf{Metrics} We employ evaluation metrics consistent with prior works~\cite{yang2024taskweave, lei2021detecting}. For QVHighlights, we report Recall@1 (R1) at Intersection over Union (IoU) thresholds of 0.5 and 0.7. Given that this dataset allows multiple matching segments per query, we also use mean average precision (mAP) as an evaluation metric, reporting scores at IoU thresholds of 0.5, 0.75, and the average over all IoUs. For Charades-STA and TACoS, we evaluate Recall@1 (R1) at IoU thresholds of 0.3, 0.5, and 0.7, along with the mean IoU (mIoU) of the top-1 predictions.

\subsection{Implementation Details}
Following previous methods~\cite{lei2021qvhighlights, sun2024trdetr, yang2024taskweave, xiao2024uvcom}, we extract video features from the primary benchmark using CLIP~\cite{radford2021clip} and SlowFast~\cite{feichtenhofer2019slowfast}, while text features are extracted using the text branch of CLIP. For additional experiments on Charades-STA, we also employ VGG~\cite{simonyan2014vgg} or I3D~\cite{carreira2017i3d} as video encoders and use GloVe~\cite{pennington2014glove} embeddings for text representation.
Our training process consists of two stages, with 40 epochs in the first stage and 100 epochs in the second stage.
In both stages, we use the AdamW optimizer with a learning rate of 1e-4 and a weight decay of 1e-4. The loss function weights are set as $\lambda_{\text{disc}} = 0.5$ and $\lambda_{\text{dill}} = 0.5$. All experiments are conducted on a single NVIDIA A40 GPU.

\subsection{Comparison with State-of-the-arts}
\textbf{QVHighlights.} As shown in Table~\ref{tab:1}, we compare our model with previous methods on both validation and test sets of QVHighlights. Compared to the previous state-of-the-art method, BAM-DETR~\cite{lee2025bamdetr}, our model achieves significant improvements of +5.03\% and +5.04\% in R1@0.5 and R1@0.7 scores on the validation set, respectively, along with a +4.05\% enhancement in average mAP. On the test set, our model maintains substantial advantages, notably surpassing BAM-DETR by +5.51\% in R1@0.5. 

\input{table/vgg_and_i3d}
\input{table/ablation}
\begin{figure*}[h]
    \centering
    \includegraphics[width=1\linewidth]{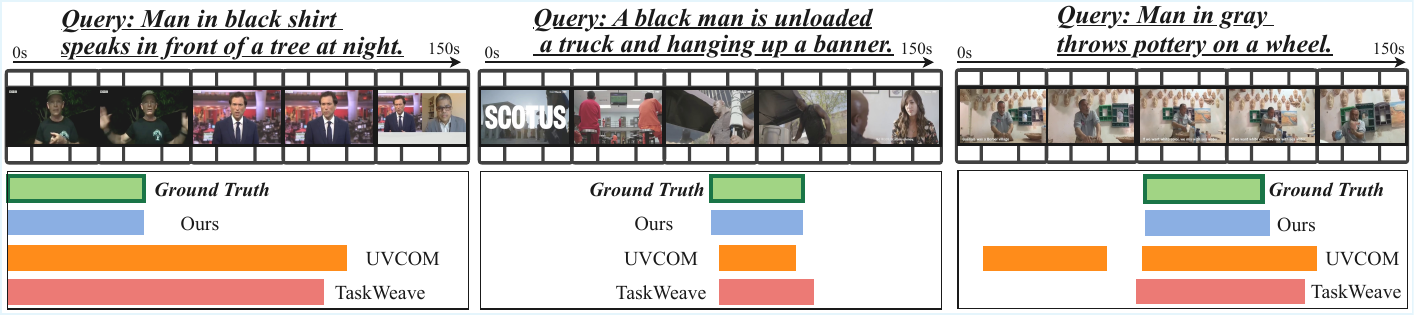}
    \caption{Qualitative comparison with other methods on QVHighlights validation set.}
    \label{fig:quali}
\end{figure*}

\noindent\textbf{Charades-STA and TACoS.} In Table~\ref{tab:2}, we further compare our model with previous methods on the Charades-STA and TACoS datasets. Additionally, Table~\ref{tab:3} presents a comparison of different models on the Charades-STA dataset when replacing the backbone. Experimental results demonstrate that our model consistently achieves the best performance.
Specifically, compared to the previous best-performing model UVCOM~\cite{xiao2024uvcom}, our model achieves an average performance gain of +3.40\% on Charades-STA and +3.64\% on TACoS across different metrics. When replacing the backbone, our model also outperforms the previous state-of-the-art TR-DETR~\cite{sun2024trdetr}, with an average improvement of +4.51\% for the VGG backbone and +3.27\% for the I3D backbone across different metrics. These results highlight the strong robustness of our approach across various video types and feature extraction methods.

\subsection{Ablation Study and Discussion}
In this section, we conduct ablation studies on the validation set of QVHighlights to further analyze the effectiveness of our proposed components. Setting (a) represents our baseline model, which is the M-DETR~\cite{lei2021qvhighlights} model combined with the local-global alignment module and visual feature refinement module proposed in \cite{sun2024trdetr}. In settings (b) and (c), we use data augmentation without the two-stage training strategy. Although the results of (c) are better than (b), both still underperform compared to the baseline model. This is due to the significant difference between the construction of augmented data and real data. If the model is directly trained on augmented data, it may overfit these synthetic samples and fail to adapt to the distribution of real data. Setting (e) shows another aspect; using only two-stage training without data augmentation (i.e., multi-round training on real data) does not improve model performance. However, when we combine data augmentation with two-stage training, model performance improves significantly, as shown in settings (e) and (f). This demonstrates that the two-stage training approach can leverage the advantages of augmented data and adapt to the distribution of real data. In settings (g) and (h), we further incorporate the dual-path distillation module into the two-stage training framework. This module can effectively distill knowledge from the first stage, further enhancing model performance. Settings (i) and (j) show the performance of adding the discriminative contrastive loss to the two-stage model. This loss improves the model's semantic understanding ability by requiring it to distinguish ambiguous fragments. Finally, settings (k) and (l) use all modules of the two-stage training. Setting (l) also uses both data augmentation strategies simultaneously, achieving the best performance across all metrics.

\subsection{Qualitative Results}
As shown in Fig.~\ref{fig:quali}, we conduct a qualitative comparison between our approach and recent open-source methods, UVCOM~\cite{xiao2024uvcom} and TaskWeave~\cite{yang2024taskweave}, on the validation split of QVHighlights.
In the first case, the query text is ``man in black shirt speaks in front of a tree at night." In the video, immediately following the ground truth segment, another man wearing black clothing appears. However, he does not satisfy the condition ``in front of a tree at night." Both UVCOM and TaskWeave incorrectly localize this ambiguous segment, whereas our model successfully distinguishes it.
In the second case, although all models identify the approximate location of the target segment, UVCOM and TaskWeave fail to precisely align its boundaries compared to our method.
In the third case, while all models roughly identify the relevant segment, UVCOM and TaskWeave fail to accurately localize the boundaries between similar actions (e.g., throwing vs. rubbing), whereas our method achieves more precise segmentation.
These results demonstrate the superiority of our model in semantic comprehension and precise localization.

%% file: table/tacos_and_cha.tex
\begin{table*}
    \centering
    \setlength\tabcolsep{3mm}
    \begin{tabular}{l|cccc|cccc}
    \toprule
    \multirow{2.5}{*}{\textbf{Method}} & \multicolumn{4}{c|}{\textbf{Charades-STA}} & \multicolumn{4}{c}{\textbf{TACoS}}\\
    \cmidrule(rl){2-5} \cmidrule(rl){6-9}
    & R1@0.3 & R1@0.5 & R1@0.7 & mIoU & R1@0.3 & R1@0.5 & R1@0.7 & mIoU\\
    \midrule
    2D-TAN~\cite{zhang2020tan} & 58.76 & 46.02 & 27.50 & 41.25 & 40.01 & 27.99 & 12.92 & 27.22 \\
    M-DETR~\cite{lei2021qvhighlights} & 65.83 & 52.07 & 30.59 & 45.54 & 37.97 & 24.67 & 11.97 & 25.49 \\
    MomentDiff~\cite{li2023momentdiff} & - & 55.57 & 32.42 & - & 44.78 & 33.68 & - & - \\
    QD-DETR~\cite{moon2023qd-detr} & - & 57.31 & 32.55 & - & - & - & - & - \\
    UniVTG~\cite{lin2023univtg} & 70.81 & 58.01 & 35.65 & 50.10 & 51.44 & 34.97 & 17.35 & 33.60 \\
    UVCOM~\cite{xiao2024uvcom} & - & 59.25 & 36.64 & - & - & 36.39 & 23.32 & - \\
    \rowcolor{gray!50} 
    \textbf{Ours} & \textbf{72.12} & \textbf{62.02} & \textbf{40.67} & \textbf{52.58} & \textbf{55.16} & \textbf{40.91} & \textbf{26.07} & \textbf{38.22} \\
    \bottomrule
    \end{tabular}
\caption{Experimental results on Charades-STA and TACoS. The video features are extracted from SlowFast and CLIP.}
\label{tab:2}
\vspace{-1mm}
\end{table*}

%% file: table/vgg_and_i3d.tex
\begin{table}
    \centering
    \setlength\tabcolsep{3.5mm}
    \renewcommand\arraystretch{1}
    \begin{tabular}{l|c|cc}
    \toprule
    \textbf{Method} & \textbf{Feat} & R@0.5 & R@0.7\\
    \midrule 
    2D-TAN~\cite{zhang2020tan} & VGG & 40.94 & 22.85 \\
    UMT~\cite{liu2022umt} & VGG & 48.31 & 29.25 \\
    MomentDiff~\cite{li2023momentdiff} & VGG & 51.94 & 28.25 \\
    QD-DETR~\cite{moon2023qd-detr} & VGG & 52.77 & 31.13 \\
    TR-DETR~\cite{sun2024trdetr} & VGG & 53.47 & 30.81 \\
    \rowcolor{gray!50} 
    \textbf{Ours} & VGG & \textbf{55.56} & \textbf{37.74} \\
    \bottomrule
    VSLNet~\cite{zhang2020vslnet} & I3D & 47.31 & 30.19 \\
    QD-DETR~\cite{moon2023qd-detr} & I3D & 50.67 & 31.02 \\
    TR-DETR~\cite{sun2024trdetr} & I3D & 55.51 & 33.66 \\
    \rowcolor{gray!50} 
    \textbf{Ours} & I3D & \textbf{58.15} & \textbf{36.96} \\
    \bottomrule
    \end{tabular}
    \caption{Experimental results on Charades-STA, the video features are extracted from VGG or I3D.}
    \vspace{-4mm}
    \label{tab:3}
\end{table} 

%% file: table/ablation.tex
\begin{table*}[h]
\centering
\setlength\tabcolsep{3mm}
\begin{tabular}{c|cc|ccc|cc|ccc}
\toprule
\multirow{2.5}{*}{\textbf{Setting}} & \multicolumn{2}{c|}{\textbf{Augment}} & \multicolumn{3}{c|}{\textbf{Two Stage}} & \multicolumn{2}{c|}{R1}& \multicolumn{3}{c}{mAP} \\
\cmidrule(rl){2-3} \cmidrule(rl){4-6} \cmidrule(rl){7-8} \cmidrule(rl){9-11}
 & \textbf{Splice} & \textbf{Boost} & \textbf{Vanilla} & \textbf{+Dill} & \textbf{+DCL} & @0.5 & @0.7 & @0.5 & @0.75 & Avg. \\
 \midrule
 (a) & - & - & - & - & - & 65.61 & 49.92 & 67.16 & 46.37 & 44.82 \\
 \midrule
 (b) & \checkmark & - & - & - & - & 63.68 & 47.29 & 63.77 & 42.65 & 42.56 \\
 (c) & \checkmark & \checkmark & - & - & - & 64.97 & 49.81 & 64.18 & 44.23 & 43.88 \\
 (d) & - & - & \checkmark & - & - & 64.58 & 48.84 & 66.27 & 45.77 & 44.66 \\
 \midrule
 (e) & \checkmark & - & \checkmark & - & - & 67.68 & 54.00 & 66.42 & 49.34 & 47.11 \\
 (f) & \checkmark & \checkmark & \checkmark & - & - & 68.52 & 54.26 & 67.84 & 49.35 & 48.07 \\
 (g) & \checkmark & - & - & \checkmark & - & 68.90 & 54.90 & 69.11 & 50.48 & 49.11 \\
 (h) & \checkmark & \checkmark & - & \checkmark & - & 69.55 & 55.55 & 69.49 & 50.32 & 50.03 \\
 (i) & \checkmark & - & - & - & \checkmark & 68.71 & 54.77 & 68.01 & 48.45 & 47.87 \\
 (j) & \checkmark & \checkmark & - & - & \checkmark & 69.23 & 54.84 & 68.14 & 49.24 & 48.80 \\
 (k) & \checkmark & - & - & \checkmark & \checkmark & 69.81 & 55.94 & 69.35 & 52.41 & 50.64 \\
 \rowcolor{gray!50} 
 (l) & \checkmark & \checkmark & - & \checkmark & \checkmark & \textbf{70.13} & \textbf{56.65} & \textbf{70.28} & \textbf{53.20} & \textbf{51.66} \\
\bottomrule
\end{tabular}
\caption{Effect of the proposed component on QVHighlights validation set. \textbf{Vanilla} means that we conduct two-stage training on real data without extra modules. \textbf{Dill} and \textbf{DCL} represent dual-path distillation module and discriminative contrastive loss module respectively.}
\label{tab:4}
\end{table*}

%% file: sec/5_conclusion.tex
\section{Conclusion}
\label{sec:conclusion}
In this paper, we propose a Splice and Boost data augmentation strategy that constructs samples with clear discriminative boundaries and semantically challenging examples. To address the limitation that existing single-stage frameworks struggle to effectively leverage both augmented and real data, we introduce a two-stage training framework with Cold-start and Dual-path Distillation. The model first acquires robust discriminative capabilities from augmented data, then generalizes to real-world scenarios while preserving learned capabilities. Experimental results demonstrate the superiority of our approach.
\clearpage
\noindent\textbf{Acknowledgment:} This work is supported by the National Natural Science Foundation of China (No.62206174). 